\pdfoutput=1

\documentclass[11pt]{article}

\usepackage{graphicx}
\usepackage{arabtex}
\usepackage{utf8}
\setcode{utf8}
\usepackage[table]{xcolor}

\usepackage{multirow}

\usepackage{tipa}
\usepackage{arabtex}
\usepackage{utf8}
\usepackage{tipa}
\usepackage{diagbox}
\usepackage{xcolor}
\usepackage{makecell}

\usepackage{appendix}
\usepackage{etoolbox}

\usepackage{xspace}
\newcommand{\Ar}[1]{{\scriptsize \<#1>\xspace}}
\newcommand{\TrAr}[1]{\arabtrue\transfalse{\scriptsize \Ar{#1}} \arabfalse\transtrue \RL{#1}\arabtrue\transfalse \hspace{-0.5ex}}

\usepackage[]{EMNLP2023}

\usepackage{times}
\usepackage{latexsym}

\usepackage[T1]{fontenc}

\usepackage[utf8]{inputenc}

\usepackage{microtype}

\usepackage{inconsolata}

\usepackage{abstract}
\title{SALMA: Arabic Sense-Annotated Corpus and WSD Benchmarks}

\author{Mustafa Jarrar, Sanad Malaysha, Tymaa Hammouda, Mohammed Khalilia \\
   Birzeit University, Palestine \\
    \texttt{\{mjarrar, smalaysha, thammouda, mkhalilia\}@birzeit.edu} \\
 }

\begin{document}
\maketitle
\begin{abstract}
SALMA, the first Arabic sense-annotated corpus, consists of \textasciitilde34K tokens, which are all sense-annotated. The corpus is annotated using two different sense inventories simultaneously (Modern and Ghani). SALMA novelty lies in how tokens and senses are associated. Instead of linking a token to only one intended sense, SALMA links a token to multiple senses and provides a score to each sense. A smart web-based annotation tool was developed to support scoring multiple senses against a given word. In addition to sense annotations, we also annotated the corpus using six types of named entities. The quality of our annotations was assessed using various metrics (Kappa, Linear Weighted Kappa, Quadratic Weighted Kappa, Mean Average Error, and Root Mean Square Error), which show very high inter-annotator agreement.
To establish a Word Sense Disambiguation baseline using our SALMA corpus, we developed an end-to-end Word Sense Disambiguation system using Target Sense Verification. We used this system to evaluate three Target Sense Verification models available in the literature. Our best model achieved an accuracy with 84.2\% using Modern and 78.7\% using Ghani. The full corpus and the annotation tool are open-source and publicly available at {\scriptsize  \url{https://sina.birzeit.edu/salma/}}.
\end{abstract}

\section{Introduction}
\label{sec:introduction}
WSD aims to determine a word's intended meaning (sense) in a given context. WSD is underdeveloped in Arabic due to the lack of sense-annotated datasets. This is in addition to the challenging nature of the WSD task due to the semantic polysemy of the words \citep{HJ21b}. For instance, the Arabic word (\TrAr{عَين}) has sixteen meanings in the Contemporary Arabic Dictionary \citep{omar2008contemporary}. In the context (\TrAr{رأيتُه رأي العَين}), word (\TrAr{عَين}) refers to \textit{eye}, while in (\TrAr{شرِبت مِن عَين الماء}), it refers to \textit{water spring}. Similarly, the English word \textit{book} as a noun has ten different senses in Princeton WordNet \citep{miller1990WordNet}, such as (a written work or composition that has been published), or (number of pages bound together). WSD has been considered a challenging task for many years \citep{weaver1952translation}, but it has recently gained more attention due to the advances in learning contextualized word representations from language models, such as BERT \citep{devlinbert} and GPT \citep{radfordgpt}. 

As glosses are short descriptions of senses \cite{J06,J05}, recent research has demonstrated promising results in WSD task by framing the problem as a sentence-pair (context-gloss) binary classification task, referred to as Target Sense Verification (TSV), where the context is a sentence containing the ambiguous word \citep{Huangwsd,yap2020adapting, blevinscontextgloss}. \citet{HJ21b} proposed an approach for Arabic WSD (using TSV) based on context-gloss pairs extracted from the Arabic Ontology and lexicons and they achieved 84\% accuracy, but this evaluation was done on a TSV dataset rather than a WSD evaluation using a sense-annotated corpus. Additionally, \citet{HJ21b} presented an attempt for Arabic Word-in-Context (WiC) disambiguation using the dataset provided by the SemEval shared task \citep{mclwic}.

This article presents SALMA, the first sense-annotated Arabic corpus consisting of about 34K tokens, which are manually annotated with senses. Since there are no available sense inventories for Arabic, We used two Arabic lexicons as sense inventories: Contemporary Arabic Dictionary (\TrAr{اللغة العربية المعاصرة}), hereafter we refer to as \textbf{Modern} \citep{omar2008contemporary}, and Al-Ghani Al-Zaher (\TrAr{الغني الزاهر}), hereafter we refer to as \textbf{Ghani} \citep{abul2014ghani}. These two lexicons are part of the lexicon digitization project and lexicographic database at SinaLab\footnote{https://sina.birzeit.edu/} \citep{JA19,ADJ19,ADJ19_report, GJJB23,JKKS21}. We introduce a novel sense-annotation framework (Section \ref{sec:annotation}), in which all candidate senses, from both lexicons, are scored to indicate their semantic relatedness to a token appearing within a context. The higher the score, the more semantically related the sense is. For better coverage, we annotated each token in our corpus using both lexicons independently and in parallel. The scores assigned to senses of the Modern do not influence the scoring of the Ghani senses. In addition, we also annotated our corpus using six types of named entities: person (PERS), organization (ORG), geopolitical entity (GPE), location (LOC), facility (FAC), and currency (CURR). The corpus was annotated by three linguists and we assessed the inter-annotator agreement (IAA) using 2.6\% of the annotated words in the corpus. 
To establish a baseline for WSD in Arabic, we developed an end-to-end WSD system, in which we benchmarked three available TSV models, with different settings. The best model resulted in 84.2\% accuracy using Modern and 78.7\% using Ghani.
The main contributions of this paper are:
\begin{itemize}
   \item \textit{Sense-annotated corpus}, annotated with two sense inventories independently, and six named entities; and most importantly, each word is linked with all of its senses, and each sense is given a score.
   \item \textit{Web-based sense-annotation framework} to score all senses of a given word.
   \item \textit{End-to-end WSD system}, implemented and evaluated using three different TSV models. 
   \item \textit{WSD baseline for Arabic}, with different settings.
\end{itemize}

The remainder of the article is organized as follows: Section \ref{sec:related-work} highlights the related work, Section \ref{sec:annotation} presents the corpus,  Section \ref{sec:iaa} describes the inter-annotator agreement, Sections \ref{sec:wsd-salma} and \ref{sec:Baselines} present how the baselines are produced, we conclude in Section \ref{sec:conclusion} and outline the limitations and future work in Section \ref{sec:limits-future}.

\section{Related Work}
\label{sec:related-work}
We will first review related sense-annotated corpora, then we will review related sense inventories.

One of the known English sense-annotated corpora is SemCor \citep{miller1993semantic}, which is annotated using the Princeton WordNet \citep{miller1990WordNet}. It contains about 200K sense annotations for around 700K words, but not all words are sense-annotated in the SemCor corpus, especially multi-word expressions, articles, and prepositions. The AnCora corpus for Spanish and Catalan languages \citep{marti2007ancora} was collected from newspapers and consists of 500K words, but only 200K noun words are semantically annotated using the Spanish WordNet. AnCora also includes morphological, semantic, and syntactic annotations. TuBa-D/Z is a German annotated corpus, manually collected from newspapers and annotated using the GermanNet senses \citep{henrich2013tueba}. TuBa-D/Z was later used as a gold standard for the WSD task by \citep{petrolitowncorporasurvey}. The  Italian Syntactic-Semantic Treebank (ISST) is a corpus built for the Italian language with 89,941 sense-annotated words \citep{montemagni2003building}. The ISST annotations cover five levels that are related to lexico-semantics such as orthographic, morpho-syntactic, semantic, and syntactic aspects.

The NTU-MC corpus \citep{Tanntumc} covers eight languages including Thai, Vietnamese, Arabic, Korean, Indonesian, Japanese, Mandarin Chinese, and English. However, the Arabic version is not publicly available. This corpus was collected from short stories, essays, and tourism articles resulting in a total of 116K words, but only 63K words are annotated. 
KPWr, a Polish corpus, contains text from multiple domains including science, law, religion, and press \citep{broda-etal-2012-kpwr} with a total of 438,327 words, but only 9,157 words are annotated using the Polish WordNet \citep{maziarz2012approaching}.

\begin{table*}[h!]
\centering
\small
\begin{tabular}{|c|c|c|c|ccccc|}
\hline
\multirow{2}{*}{\textbf{Corpus}} &
  \multirow{2}{*}[-1.5ex]{\textbf{\begin{tabular}[c]{@{}c@{}}Unique\\ Senses\end{tabular}}} &
  \multirow{2}{*}[-1.5ex]{\textbf{\begin{tabular}[c]{@{}c@{}}Annotation\\ Type\end{tabular}}} &
  \multirow{2}{*}[-1.5ex]{\textbf{\begin{tabular}[c]{@{}c@{}}Corpus\\ Size \\ [-1.1ex]\scriptsize(tokens)\end{tabular}}} &
  \multicolumn{5}{c|}{\textbf{Annotations}} \\ \cline{5-9} 
 &
   &
   &
   & 
  \multicolumn{1}{c|}{Nouns} &
  \multicolumn{1}{c|}{Verbs} &
  \multicolumn{1}{c|}{\begin{tabular}[c]{@{}c@{}}Func.\\ Words\end{tabular}} &
  \multicolumn{1}{c|}{\begin{tabular}[c]{@{}c@{}}Punc.+\\ Digits\end{tabular}} &
  Total \\ \hline
AQMAR &
  \begin{tabular}[c]{@{}c@{}}25 semantic fields\\ \scriptsize(closer to named entities)\end{tabular}
  &
  \begin{tabular}[c]{@{}c@{}}selected words\\ each one sense\end{tabular} &
  \multicolumn{1}{|r|}{65K} &
  \multicolumn{1}{r|}{\textasciitilde22K} &
  \multicolumn{1}{r|}{–} &
  \multicolumn{1}{r|}{–} &
  \multicolumn{1}{r|}{–} &
  \multicolumn{1}{r|}{\textasciitilde22K} \\ \hline
OntoNotes5 & 
  \begin{tabular}[c]{@{}c@{}}261 semantic fields\\ \scriptsize(high-level grouped senses)\end{tabular}  
  &
  \begin{tabular}[c]{@{}c@{}}selected words\\ each one sense\end{tabular} &
  \multicolumn{1}{r|}{300K} &
  \multicolumn{1}{r|}{8,700} &
  \multicolumn{1}{r|}{4,300} &
  \multicolumn{1}{r|}{–} &
  \multicolumn{1}{r|}{–} &
  \multicolumn{1}{r|}{13K} \\ \hline
\begin{tabular}[c]{@{}c@{}}\textbf{SALMA}\\ \textbf{(ours)}\end{tabular} &
  \begin{tabular}[c]{@{}c@{}}4,151 word senses \\\scriptsize(from each sense inventory) \\ 6 types of named entities\end{tabular} & \begin{tabular}[c]{@{}c@{}}all senses of\\all words\end{tabular} 
  &
  \multicolumn{1}{r|}{34K} &
  \multicolumn{1}{r|}{\textbf{19,030}} &
  \multicolumn{1}{r|}{\textbf{2,763}} &
  \multicolumn{1}{r|}{\textbf{7,116}} &
  \multicolumn{1}{r|}{\textbf{5,344}} &
  \multicolumn{1}{r|}{\textbf{34,253}} \\ \hline
\end{tabular}
\caption{\label{corpus-comparison}Overview of related Arabic sense-annotated corpora.}
\end{table*}

For Arabic, the focus of research has been primarily on developing corpora for morphological and syntactic tagging \cite{DH21} rather than semantic and sense annotation, as noted by \citet{Elayeb19} and \citet{KAJ21}. For instance, part of the OntoNotes corpus \citep{ontonotes} covers limited semantic annotations for Arabic using a small sense inventory of size 261 senses (150 verbs and 111 nouns). Additionally, AQMAR corpus \citep{SchneiderMOS12} is annotated with 25 super-sense labels representing broad semantic fields such as ARTIFACT and PERSON, which can be considered as general types of named entities, rather than word-sense annotations. They annotated \textasciitilde22K nouns out of 65K tokens corpus. Table \ref{corpus-comparison}  compares our proposed corpus and related Arabic resources. 

In addition to the lack of sense-annotated corpora, Arabic lacks reliable sense inventories. Although there are some available semantic resources, they are not mature enough to be used as sense inventories. For example, the Arabic WordNet \citep{black2006introducing} contains about 10K senses, and the Arabic Ontology \citep{J21,J11} contains about 18K synsets. However, both resources cannot be used as sense inventories as they do not provide a complete set of senses for a given lemma (i.e., lexicon entry). The lexicographic database developed at Birzeit University contains about 150 Arabic lexicons \citep{JA19,JAM19}, but these lexicons are not well-structured or suitable to be used as sense inventories \citep{JA19}. Due to the lack of dependable  Arabic sense inventory, we decided to obtain a license to digitize and use two Arabic lexicons as sense inventories, namely, \textbf{Modern} \citep{omar2008contemporary} and \textbf{Ghani} \citep{abul2014ghani}.

\section{Corpus Construction and Annotation}
\label{sec:annotation}

\subsection{Corpus Collection}
\label{sec:collection}
Our SALMA corpus is part of the Wojood corpus \cite{JKG22}, and was collected from 33 online media sources written in Modern Standard Arabic (MSA) and covering general topics. Some of those sources include \href{https://mipa.institute/}{mipa.institute}, \href{https://sanaacenter.org/}{sanaacenter.org}, \href{https://www.hrw.org/}{hrw.org}, \href{https://www.diplomatie.ma/ar}{diplomatie.ma}, \href{https://sa.usembassy.gov/}{sa.usembassy.gov}, \href{https://www.eeas.europa.eu/_en}{eeas.europa.eu}, \href{https://www.crisisgroup.org/ar}{crisisgroup.org}, and \href{https://www.mofaic.gov.ae/ar-ae/}{mofaic.gov.ae}. The corpus was then segmented into sentences and tokenized, resulting in 1439 sentences and \textasciitilde34K tokens, with an average of 23.8 tokens per sentence.


\subsection{Annotation Framework}
\label{sec:guidelines}
This section presents a novel sense annotation framework, where instead of linking a word to one sense, we propose to score all semantically related senses to the word. The score ranges between 1-100\% and a sense with a score {\small $\geq60\%$} is considered a correct sense of the word. The ranking scale is divided into six categories:

\begin{itemize}
    \item \textit{Explicate} /\Ar{مباشرة} (100\%): direct and explicate semantics (\Ar{دلالة صحيحة وصريحة}).
    \item \textit{General} /\Ar{معنى عـام} (80\%): correct but implicate semantics (\Ar{دلالة صحيحة غير مباشرة}).
    \item \textit{Referral} /\Ar{دلالة لغوية} (60\%): generally correct semantics, but is referred to another lemma (\Ar{صحيحة ولكن عامة جداً مثل مصدر، اسم فاعل}). For example, the word \textit{drinker} and its gloss (\textit{active participle of drink}).
     \item \textit{Related} /\Ar{ذات علاقة} (40\%): weak semantics (\Ar{مشتركة في الدلالة العامة فقط، أختها دلالياً}). For example, the term (\TrAr{سياسة الشركة}) / \textit{company's policy}, is related to the sense (\textit{the policy used to collect taxes}) which is not a sense of the lemma (\TrAr{سياسة}), but semantically related.
     \item \textit{Root semantics} /\Ar{دلالة جذر} (20\%): share root semantics (\Ar{دلالة مختلفة ولكن تشترك في الدلالة المجردة التي يحملها \\ الجذر، مثل الدلالة المجازية}). In Arabic lexical semantics, all words with the same root share part of the semantics of this root \citep{ryding2014arabic, boudelaa2004abstract, boudelaa2010arabic}. For example, all senses of the lemma (\TrAr{سياسة}), such as \textit{politics} and
     \textit{policies} share an abstract meaning (e.g., issues related to governing and acting).
     \item \textit{Different} /\Ar{مخـتـلفـــة}  (1\%): unrelated semantics (\Ar{دلالة مختلفة تماماً}).
\end{itemize}

\begin{figure*}[ht!]
    \centering    \includegraphics[width=1\textwidth,height=11cm]{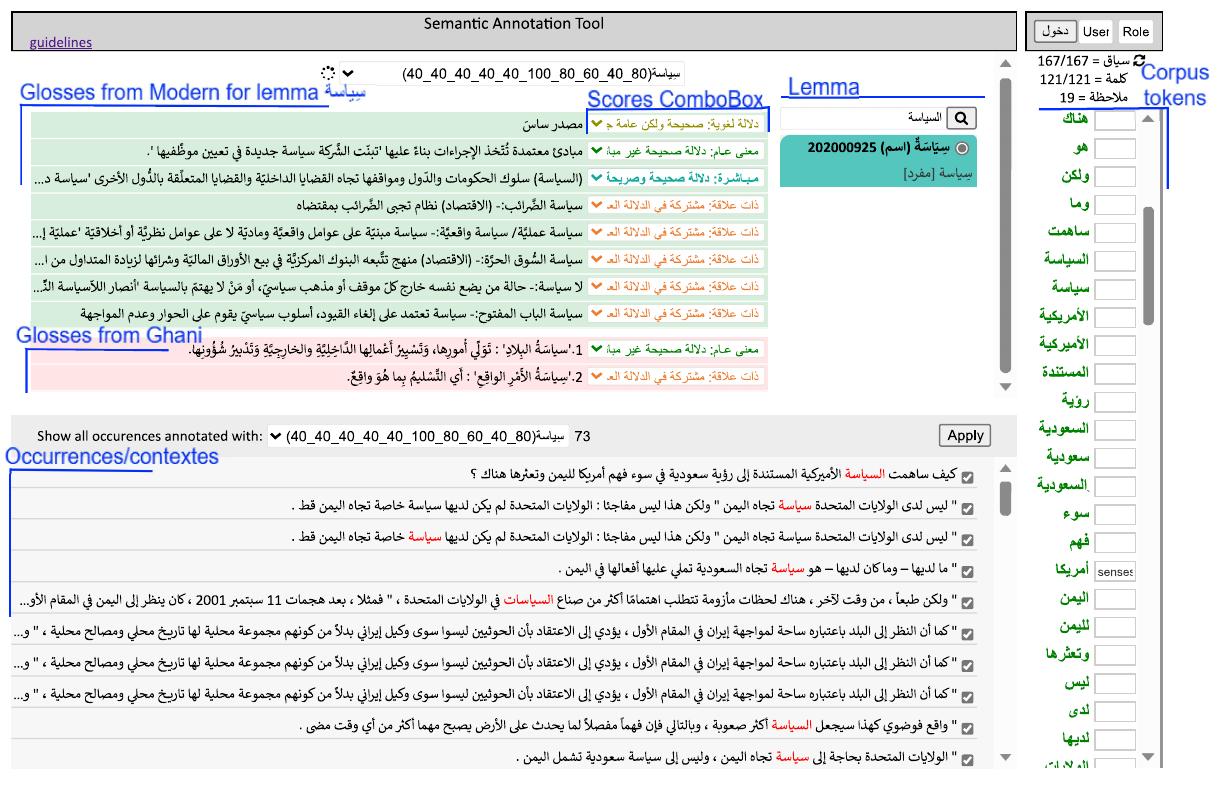}
    \caption{Screenshot of our web-based annotation tool.}
    \label{fig:Sense_annotated}
\end{figure*}

This framework serves several purposes. First, in case of underdeveloped sense inventories (such as the Modern and Ghani lexicons), in which glosses might be vague, redundant, or overlapping, our framework allows the annotators to score each sense. In this paper, we linked every word in the corpus with all semantically related senses in Modern and Ghani, thus we were able to compare and evaluate the lexical coverage in both lexicons (see Section \ref{sec:Lexical_Coverage}). Another advantage of using this framework (i.e., scoring all senses) is that our corpus can be used to benchmark ranking-based WSD methods \cite{Coniawsdmultilabel, yap2020adapting}, which is not possible in the case of one-sense annotated corpora.

\subsection{Annotation Tool} \label{sec:annotation_tool}
We developed a web-based tool optimized for our sense annotation framework and methodology. On the right side of Figure \ref{fig:Sense_annotated}, the linguist selects a word to be annotated (such as "\TrAr{السياسة}"). The tool will then retrieve all sentences (i.e. contexts) in the corpus containing the selected word. The tool will also automatically fetch the lemma of the selected word, and the linguist has the ability to search for the lemma manually. After selecting a lemma, the tool retrieves senses associated with the lemma from both lexicons, Modern and Ghani. The linguist can then select the score category for each sense according to our guideline and apply these scores to all selected words (in contexts) as shown in Figure \ref{fig:Sense_annotated}. The scores are selected from a \texttt{ComboBox} of the six categories (See Section \ref{sec:guidelines}), however, the tool internally stores their corresponding numeric values.

\subsection{Annotation Process}
\label{sec:Annotation_Process}

The annotation was carried out in three phases: 

\noindent\textbf{Phase 1 (training)}: we recruited three undergraduate students majoring in linguistics. The students were trained in three steps in order to produce consistent annotations. We first assigned 50 words to each linguist and trained them to conduct the annotation jointly. Second, we assigned the same 150 words to each student separately, then asked them to compare and consolidate their annotations, which helps in calibrating their scoring. Third, we repeated the second phase, but using 300 words and again we asked them to compare their annotations.

\noindent\textbf{Phase 2 (annotation}): out of \textasciitilde34K tokens, excluding digits and punctuations, we assigned about 9.6K words to each of the three linguists. Each linguist was asked to annotate all occurrences of each word in the corpus - resulting in about \textasciitilde29K annotations for the whole words.
  
\noindent\textbf{Phase 3 (validation}): after finishing the annotations, we used the tool to automatically validate the annotations and flag those that violated the following cases: (i) a word is annotated with more than one \textit{Explicit} or \textit{General} sense in the same lexicon, which is an indication of either a mistake or redundant or overlapping senses in the lexicon. (ii) a word is missing either an \textit{Explicit} or a \textit{General} sense; this is an indication of a mistake or the lexicon is missing this sense. (iii) if the selected sense is a proper noun, then all other senses should be ranked as \textit{Different}. The linguists were asked to review these flagged annotations and revise them if necessary.

The linguists were encouraged to discuss among themselves and take joint decisions when facing difficulties, especially in the case of vague glosses or contexts. In addition, as will be discussed in Section \ref{sec:Lexical_Coverage}, missing lemmas and senses are manually added to the lexicons. Table \ref{table:SALMA-details} provides general statistics about the annotations. It is worth noting that sense annotations are typically costly and time-consuming. The linguists spent about 600 working days (i.e., 4800 working hours) to carry out the three phases described above.

\begin{table}[ht!]
\small 
\centering
\begin{tabular}{|p{0.95cm}|l|l|l|l|l|}
\hline
  \textbf{Term} &
  {\textbf{Noun}} &
  {\textbf{Verb}} &
  {\textbf{\makecell[l]{Func.\\ Words}}} &
  {\textbf{\makecell[l]{Punc+\\Digits}}} &
  \textbf{Total} \\ \hline
 \multicolumn{1}{|l|}{Tokens}                                                  & \multicolumn{1}{r|}{19,030}  & \multicolumn{1}{r|}{2,763} & \multicolumn{1}{r|}{7,116} & \multicolumn{1}{r|}{5,344} & \multicolumn{1}{r|}{\textbf{34,253}} \\ \hline
\begin{tabular}[c]{@{}l@{}}Unique\\ Tokens\end{tabular} & \multicolumn{1}{r|}{6,670}  & 1,593 & \multicolumn{1}{r|}{322}   & \multicolumn{1}{r|}{175}   & \multicolumn{1}{r|}{\textbf{8,760}}  \\ \hline
\begin{tabular}[c]{@{}l@{}}Unique\\  Lemmas\end{tabular} & \multicolumn{1}{r|}{2,904}  & \multicolumn{1}{r|}{677}   & \multicolumn{1}{r|}{119}   & \multicolumn{1}{r|}{175}   & \multicolumn{1}{r|}{\textbf{3,875}}  \\ \hline
\begin{tabular}[c]{@{}l@{}}Unique\\  Senses\end{tabular} & \multicolumn{1}{r|}{3,151}  & \multicolumn{1}{r|}{792}   & \multicolumn{1}{r|}{206}   & \multicolumn{1}{|r|}{2}     & \textbf{4,151}  \\ \hline
\end{tabular}
\caption{Statistics of the SALMA corpus.}
\label{table:SALMA-details}
\end{table}

\begin{table}[ht!]
\small
\centering
\begin{tabular}{|c|c|c|}
\hline
           \textbf{Term} & \textbf{Modern} & \textbf{Ghani} \\ \hline
Lemmas       & \multicolumn{1}{l|}{\small{80\%} \scriptsize(2,788/3,522)}            & \multicolumn{1}{l|}{\small{78\%} \scriptsize(2,724/3,522)}           \\ \hline
\makecell{Senses\\ \tiny (Without Proper nouns)}       & \multicolumn{1}{l|}{\small{83\%} \scriptsize(3,430/4,151)}           & \multicolumn{1}{l|}{\small{78\%} 
 \scriptsize(3,226/4,151)}            \\ \hline
\makecell{Proper Nouns\\ Senses} & \multicolumn{1}{l|}{\small{4\%} 
 \scriptsize(9/213)}            & \multicolumn{1}{l|}{\small{14\%} \scriptsize(30/213)}            \\ \hline
\end{tabular}
\caption{\label{lexicons-evaluation}Coverage of Modern and Ghani lexicons.}
\end{table} 

\subsection{Discussion and Lexical Coverage}
\label{sec:Lexical_Coverage}
We evaluated the coverage of both lexicons based on the sense-annotated tokens. As Table \ref{lexicons-evaluation} shows, Modern has higher coverage of lemmas (80\%) compared to Ghani's coverage (78\%), and has higher sense coverage (83\%) compared to Ghani (78\%). Moreover, glosses in Modern are more precise, less ambiguous and well-formulated  as discussed in Section \ref{sub:iaa-results}. The proper nouns are the main reason for the missing lemmas and senses, as the Modern and Ghani cover 4\% and 14\% of proper nouns in SALMA corpus, respectively.
Lemmas and senses that are not covered by any of the two lexicons were added manually by the linguists.
All numerical values are annotated with the same "digit" sense that covers ordinal and nominal numbers, and similarly, punctuation marks are all annotated with "Punc".


\subsection{Named Entity Annotations}
\label{sub:ner}
Named-entity annotations are important in sense-annotated corpora because sense inventories do not typically cover names of organizations, towns, people, landmarks, and others. 

\begin{table}[ht!]
\small
\centering
\begin{tabular}{|l|p{6cm}|}
\hline
  \textbf{Tag}   &  \textbf{Description} \\ \hline
PERS & Person names: first, middle, last, nickname ... \\ \hline
ORG  & Organizations: company, team, government ...\\ \hline
GPE & Geopolitical entities: country, city, state ... \\ \hline
LOC & Geographical locations: river, sea, mountain... \\ \hline
FAC & facilities: landmark, road, building, airport ...\\ \hline
CURR & Currency names or symbols.\\ \hline
\end{tabular}
\vspace*{-3mm}
\caption{\label{NER_annotation} Types of named entities.}
\end{table} 

In addition to word-sense annotations, we annotated our corpus using six types of named entities listed in Table \ref{NER_annotation}. As our corpus is a part of the Wojood, which is annotated with 21 types of nested named-entities \citep{JKG22}, in this article we annotated SALMA with six flat entities only. We used the IOB2 tagging scheme \citep{IOB2}, where B indicates the beginning of the entity mention, I the inside token, and O outside token. 

\begin{table}[ht!]
\small
\centering
\begin{tabular}{|l|r|r|}
\hline
  \textbf{Tag}   &  \textbf{\makecell[l]{Named  Entity \\Mentions}} &  \textbf{\makecell[l]{Tokens in the \\Entity  Mentions}}\\ \hline
PERS & 294 &  568 \\ \hline
ORG  & 1,123 & 2,108 \\ \hline
GPE & 1,086 & 1,295 \\ \hline
LOC & 166 & 318  \\ \hline
FAC & 22 & 59 \\ \hline
CURR & 37 & 41 \\ \hline
 \textbf{Total} &  \textbf{2,728} &  \textbf{4,389} \\ \hline
\end{tabular}
\vspace*{-2mm}
\caption{\label{NER_Stat} Statistics of named entities in SALMA corpus.}
\end{table} 

We applied the NER guidelines that were used to annotate the OntoNotes5 corpus \cite{weischedel2011ontonotes}. Table \ref{NER_Stat} presents statistics about all named entities in the SALMA corpus, which shows that 4389 (about 15\%) of the tokens are part of an entity mention. 

\section{Inter-Annotation Agreement (IAA)}
\label{sec:iaa} 
To evaluate our annotations,  
we selected 250 annotated words from each annotator $A \in \{A_1, A_2, A_3\}$,  and assigned them to a different annotator to perform double annotations. This yielded a total of 750 words (2.6\% of the annotated words) divided among three pairs of annotators, $\{(A_1, A_2), (A_1, A_3), (A_2, A_3)\}$.
Because our sense annotations contain scores (i.e., not discrete values), computing IAA is not straightforward. We chose to use various evaluation metrics especially those that take ranking into consideration. The IAA metrics used are: (i) Kappa, (ii) Linear Weighted Kappa (LWK), (ii) Quadratic Weighted Kappa (QWK), (iv) Mean Average Error (MAE), and (v) Root Mean Square Error (RMSE). 

Kappa is usually used when the data is nominal \citep{eugenio2004kappa}, so we set a threshold on the score ($\geq$60\%) in the six categories to be able to calculate Cohen's Kappa. The senses with scores above or equal this threshold carry the intended meanings that map with the context of the targeted word (See section \ref{sec:guidelines}). Nonetheless, a more suitable metric for ranked labels is either the LWK or QWK, as specified in the following equations, which we adopt from \citep{vanbelle2016newkappa}:
\begin{equation}
\label{eqn:qwk}
QWK = 1- \frac{\sum\limits_{i,j=1}^{K} \frac{(y_{i}-y_{j})^2}{(K-1)^2}.fo_{ij}}{\sum\limits_{i,j=1}^{K} \frac{(y_{i}-y_{j})^2}{(K-1)^2}.fe_{ij}}
\end{equation}

\begin{equation}
\label{eqn:qwk}
LWK = 1- \frac{\sum\limits_{i,j=1}^{K} \frac{|y_{i}-y_{j})|}{(K-1)}.fo_{ij}}{\sum\limits_{i,j=1}^{K} \frac{|y_{i}-y_{j})|}{(K-1)}.fe_{ij}}
\end{equation}

where $fo_{ij}$ is the observed frequency of the categories ($i$ and $j$) per the annotators selection, $fe_{ij}$ is the expected frequency for both annotators' selected categories, $(y_{i}-y_{jx})$  denotes the distance between the categories, and $K$ is number of categories.

Both LWK and QWK take the distance between categories into consideration, where the distance is defined as the number of categories separating the two annotators' selection. The difference is that LWK calculates the distance linearly while QWK calculates it quadratically. For measuring the ranking error deviation among annotators we used MAE and RMSE.

\begin{table}[ht!]
\small
\centering
\begin{tabular}{|m{1.5cm}|m{1.5cm}|m{2.5cm}|}
\hline
\multirow{2}{*}{\textbf{Metric}} & \multirow{2}{*}{\textbf{Lexicons}}                     & \multirow{2}{*}{\textbf{Average (STD)}}                \\
                                 &                                                        &                                                       \\ \hline
Kappa                            & \begin{tabular}[c]{@{}c@{}}Modern\\ Ghani\end{tabular} & \begin{tabular}[c]{@{}c@{}}90.48 ($\pm2.97$)\\ 78.68 ($\pm8.49$)\end{tabular} \\ \hline
\begin{tabular}[c]{@{}l@{}}LWK\end{tabular} &
  \begin{tabular}[c]{@{}c@{}}Modern\\ Ghani\end{tabular} &
  \begin{tabular}[c]{@{}c@{}}88.29 ($\pm5.37$)\\ 79.56 ($\pm9.35$)\end{tabular} \\ \hline
\begin{tabular}[c]{@{}l@{}}QWK\end{tabular} &
  \begin{tabular}[c]{@{}c@{}}Modern\\ Ghani\end{tabular} &
  \begin{tabular}[c]{@{}c@{}}91.94 ($\pm3.42$)\\ 86.03 ($\pm5.41$)\end{tabular} \\ \hline
RMSE                             & \begin{tabular}[c]{@{}c@{}}Modern\\ Ghani\end{tabular} & \begin{tabular}[c]{@{}c@{}}13.44 ($\pm3.08$)\\ 19.12 ($\pm3.06$)\end{tabular} \\ \hline
MAE                              & \begin{tabular}[c]{@{}c@{}}Modern\\ Ghani\end{tabular} & \begin{tabular}[c]{@{}c@{}}4.46 ($\pm2.04$)\\ 8.27 ($\pm3.52$)\end{tabular} \\ \hline
\end{tabular}
\caption{\label{evaluation-IAA}Inter-Annotator Agreement (IAA) average among the three linguists using different metrics.}
\end{table}

\subsection{IAA Results}
\label{sub:iaa-results}

Table \ref{evaluation-IAA} summarizes the result of the inter-annotator-agreement, the value in parenthesis is the standard deviation among pairs of annotators. Overall, we see higher agreement among the annotators for the Modern. The higher agreement is clear from all IAA metrics and the standard deviation. We see less confidence in the Ghani annotations as the IAA dropped across all metrics with higher variability among annotators, presented in higher standard deviation. Kappa was affected the most with a drop of 11.8\% when measured on the Ghani, followed by LWK with a drop of 8.73\%. QWK has the smallest drop of 5.91\% and also has the least variability among annotators. We believe the reason for the higher IAA on Modern is because Modern has better quality glosses compared to the Ghani, which has shorter glosses and in many cases are ambiguous. However, regardless of the lexicon used, we observed higher agreement among annotators as measured by LWK and QWK since they take advantage of the scores assigned to each gloss, while Kappa ignores the scoring information.  
\begin{figure}[ht!]
    \centering
    \includegraphics[scale=0.26]{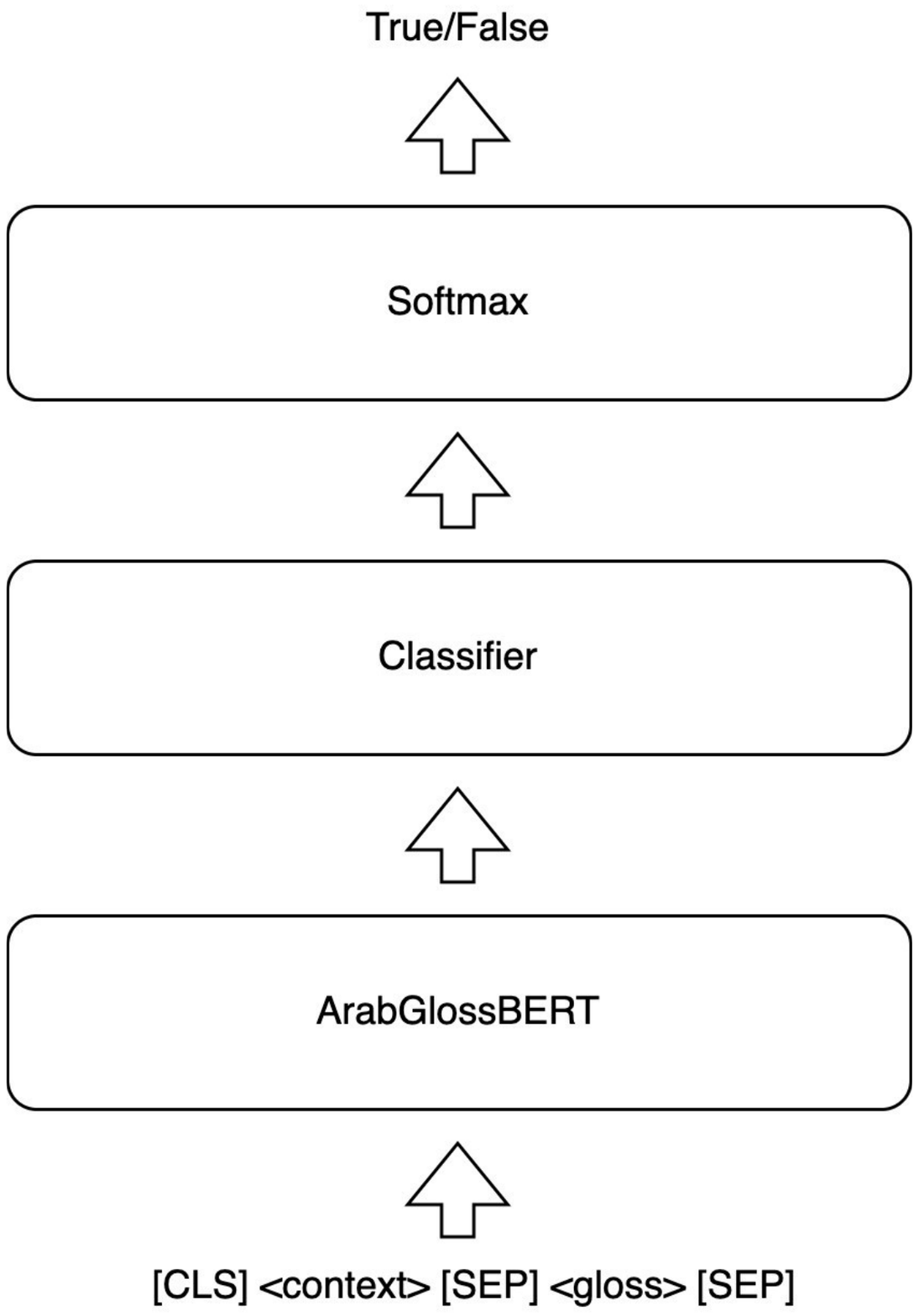}
    \caption{BERT-based TSV Architecture.}
    \label{fig:arabglossbert}
\end{figure}

\begin{figure*}[ht!]
    \centering
    \includegraphics[scale=1]{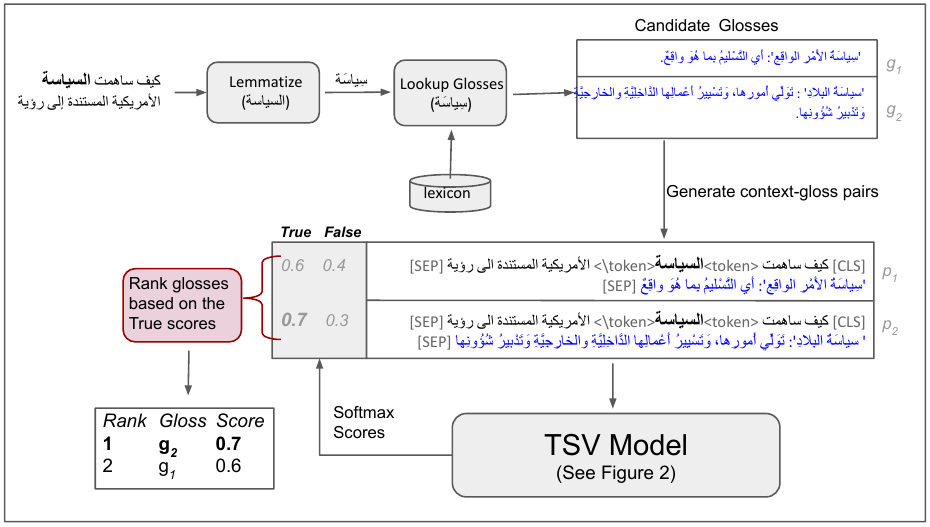}
    \caption{An end-to-end WSD using the TSV model (SALMA system).}
    \label{fig:wsd_system}
\end{figure*}

We reach similar conclusions for RMSE and MAE. Both metrics are lower for Modern compared to Ghani. The Average RMSE among all annotator pairs on the Modern is 13.44  compared to 19.12 for Ghani, while the average MAE for the Modern is 4.46 compared to 8.27 on the Ghani.

 

\section{Computing WSD Baselines using SALMA}
\label{sec:wsd-salma}
In this section, we present the baseline for Arabic WSD using our SALMA corpus. To the best of our knowledge, there are no available Arabic WSD systems to evaluate. The only available Arabic models are TSV, which are related, but not the same as WSD. In what follows, we explain the difference between WSD and TSV tasks, and propose an end-to-end WSD system using TSV. 

\subsection{The TSV Task}
\label{sec:tsv}
The TSV task is a binary classification task used to determine whether a pair of sentences (context and gloss) are True or False (see Figure \ref{fig:arabglossbert}). In other words, given a context $c$ containig the target word $w$, and a gloss $g_i$, TSV aims to classify the context-gloss pair $(c, g_i)$ as True or False. It is True if the gloss $g_i$ is the intended sense of $w$ in $c$, otherwise, it is False \citep{breit2020wic}. It is important to note that TSV is different from WSD, which determines which gloss, among a set of glosses, is the intended meaning for the target word. 

There are three available Arabic TSV models with the same architecture: (1) the Razzaz model, trained using 31K context-gloss pairs extracted from Modern \cite{el2021arabic}; (2) the ArabGlossBERT model, trained on a larger dataset (167K context-gloss pairs) extracted from several Arabic lexicons \cite{HJ21b}; and (3) the Aug-ArabGlossBERT (D9) model, trained on an augmented data, generated using back-translation of the ArabGlossBERT dataset \cite{MJK23}. 

In what follows, we propose to develop an end-to-end WSD system using TSV (called SALMA system) and in Section \ref{sec:Baselines}, we benchmark our proposed system using the SALMA corpus.

\subsection{Building WSD System Using TSV}
In this section, we propose an end-to-end solution for WSD using TSV. The solution consists of the following phases (Figure
\ref{fig:wsd_system}): 1) candidate glosses lookup, 2) target sense verification, and 3) gloss ranking.\\

\noindent\textbf{1. Candidate Glosses Lookup}: given a target word $w$ in a context $c$, we first lemmatize $w$ (i.e., determine its lemma $l$), where we use our own in-house lemmatizer, then retrieve the set of $n$ candidate glosses, $G=\{g_1, g_2,..., g_n\}$, of $l$ from the lexicon (i.e., sense inventory).

\noindent\textbf{Example}: the word $w$ (\TrAr{السياسة}) in $c$ (\Ar{كیف ساھمت السیاسة الأمریكیة المستندة الى رؤية}) has the lemma (\TrAr{سِيَاسَةٌ}) with two corresponding glosses ($\{g_1, g_2\}$) in the Ghani, as shown in Figure \ref{fig:wsd_system}.\\


\noindent\textbf{2. TSV}: once we have the set of $n$ candidate glosses, we input to the TSV model a set of $n$ context-gloss pairs, $P=\{(c, g_{i}) | \forall g_{i} \in G\}$, as illustrated with $(p_{1}, p_{2})$ in Figure \ref{fig:wsd_system}. The target word $w$ in $c$ is wrapped with special tokens "<token>$w$</token>", to emphasize the target word during training and testing of the TSV models. For each context-gloss pair, the TSV model returns confidence scores for the True and False labels, but the TSV model does not compare or rank glosses in this phase. \\

\noindent\textbf{3. Gloss Ranking}: we determine the intended meaning by ranking the glosses based on their True confidence scores calculated in the previous step. The gloss with the highest score is selected as the intended gloss for $w$.



\section{Experiments and Results}
\label{sec:Baselines}
\subsection{Experimental Setup}
To evaluate the three available Arabic TSV models using our SALMA corpus, we implemented three instances of the WSD system depicted in Figure \ref{fig:wsd_system}, each with a different TSV model. For each word in each context in the SALMA corpus, we generated context-gloss pairs similar to the example shown in Figure \ref{fig:wsd_system}. Because our corpus was sense-annotated using two lexicons (i.e., two sense inventories), we generated two sets of context-gloss pairs. In this way, we compute a separate baseline for each of the Modern and Ghani. We neither included annotations of digits and punctuations, nor the named-entity annotations presented in Section \ref{sub:ner}. 

The length of the contexts may impact the WSD accuracy, so in addition to using the full context around $w$, we also experimented with different context sizes, $s \in \{3, 5, 7, 9, 11\}$. For example, the context size $s=5$ means that there are two tokens before and two tokens after $w$.

As will be discussed in the next subsection, we evaluated three TSV models: Razzaz\footnote{We reproduced the TSV model using the code and data available at https://github.com/MElrazzaz/Arabic-word-sense-disambiguation-bench-mark}, ArabGlossBERT\footnote{ArabGlossBERT fine-tuned model  Version 1 (CC-BY-4.0) at https://huggingface.co/SinaLab/ArabGlossBERT/tree/main}, and Aug-ArabGlossBERT(D9)\footnote{Fine-tuned model D9 (CC-BY-4.0) at https://huggingface.co/SinaLab/ArabGlossBERT/tree/Augment}. We used context size $s=11$, which gave the best results. Following the authors of these models, we did not use any signal to mark up target words in the case of the Razzaz and Aug-ArabGlossBERT(D9); however, we used UNUSED0 for ArabGlossBERT. 

The experiments have been implemented in Python, specifically using the Transformers library provided by HuggigFace\footnote{https://huggingface.co/docs/transformers/index}, which is used to load and test the models. To speed-up the models evaluation, we have run the codes using a GPU (SVGA II) instance, where each run took around 20 hours. 

\begin{table}[ht!]
\small
\centering
\begin{tabular}{|m{3.5cm}|m{1.2cm}|m{1.5cm}|}
\hline
\multirow{2}{*}{\textbf{TSV Model}} & \multirow{2}{*}{\textbf{Lexicons}}& \multirow{2}{*}{\textbf{Accuracy}}\\
 & &   \\ \hline
 Razzaz  & \begin{tabular}[c]{@{}c@{}}Modern\\ Ghani\end{tabular} & 
 \begin{tabular}[c]{@{}c@{}}66.0\% \\ 68.4\% \end{tabular} \\ \hline
ArabGlossBERT & \begin{tabular}[c]{@{}c@{}}Modern\\ Ghani\end{tabular} &
  \begin{tabular}[c]{@{}c@{}}\textbf{84.2\%} \\ 77.6\% \end{tabular} \\ \hline
Aug-ArabGlossBERT(D9) & \begin{tabular}[c]{@{}c@{}}Modern\\ Ghani\end{tabular} &
  \begin{tabular}[c]{@{}c@{}}82.6\% \\ 78.7\%\end{tabular} \\ \hline
  
\end{tabular}
\caption{\label{models_Evaluations} WSD baselines for three TSV models, with context length = 11.}
\end{table}

\subsection{Baselines and Discussion}
Table \ref{models_Evaluations} presents our evaluation of the three TSV models using both Modern and Ghani with context size $s=11$. As shown in this table, the ArabGlossBERT is the best-performing model(84.2\%), which most probably because it was trained on a larger and higher quality dataset of lexicon definitions. The accuracy was calculated for nouns and verbs. We excluded the functional words as they mostly do not carry semantics.

\begin{table}[ht!]
\small
\centering
\setlength\tabcolsep{3pt}
\begin{tabular}{|c|l|lll||lll|}
\hline
\multicolumn{1}{|l|}{\multirow{2}{*}{{\textbf{Window}}}} &
  \multirow{2}{*}{{\textbf{Lexicon}}} &
  \multicolumn{3}{c||}{\textbf{\begin{tabular}[c]{@{}c@{}} Accuracy  \\ \scriptsize Target Sense Rank\end{tabular}}} &
  \multicolumn{3}{c|}{\textbf{\begin{tabular}[c]{@{}c@{}}Accuracy (Top1) \\ \scriptsize per POS \end{tabular}}} \\ \cline{3-8} 
\multicolumn{1}{|l|}{} &
   &
  \multicolumn{1}{l|}{{\scriptsize \textbf{Top1}}} &
  \multicolumn{1}{l|}{{\scriptsize \textbf{Top2}}} &
  {\scriptsize \textbf{Top3}} &
  \multicolumn{1}{l|}{{\scriptsize \textbf{Noun}}} &
  \multicolumn{1}{l|}{{\scriptsize \textbf{Verb}}} & \multicolumn{1}{l|}{{\scriptsize \textbf{Func.}}} \\ \hline
\multirow{2}{*}{All} & Modern         & \multicolumn{1}{l|}{82.8}          & \multicolumn{1}{l|}{94.2} & 97.4 & \multicolumn{1}{l|}{83.5} & \multicolumn{1}{l|}{77.9} & 41.2 \\ \cline{2-8} 
                     & Ghani          & \multicolumn{1}{l|}{77.0}          & \multicolumn{1}{l|}{89.3} & 94.1 & \multicolumn{1}{l|}{78.5} & \multicolumn{1}{l|}{66.0} & 36.0 \\ \hline
\multirow{2}{*}{11}  & \textbf{Modern}         & \multicolumn{1}{l|}{\textbf{84.2}}          & \multicolumn{1}{l|}{95.1} & 98.1 & \multicolumn{1}{l|}{85.4} & \multicolumn{1}{l|}{76.1} & 37.9 \\ \cline{2-8} 
                     & \textbf{Ghani}          & \multicolumn{1}{l|}{\textbf{77.6}}          & \multicolumn{1}{l|}{90.1} & 94.9 & \multicolumn{1}{l|}{79.4} & \multicolumn{1}{l|}{61.7} & 31.8 \\ \hline
\multirow{2}{*}{9}   & Modern         & \multicolumn{1}{l|}{83.5}          & \multicolumn{1}{l|}{95.0} & 97.9 & \multicolumn{1}{l|}{84.4} & \multicolumn{1}{l|}{78.3} & 37.7 \\ \cline{2-8} 
                     & GHani          & \multicolumn{1}{l|}{77.3}          & \multicolumn{1}{l|}{90.1} & 94.8 & \multicolumn{1}{l|}{79} & \multicolumn{1}{l|}{63.7} & 32.2 \\ \hline
\multirow{2}{*}{7}   & Modern         & \multicolumn{1}{l|}{83.8}          & \multicolumn{1}{l|}{95.1} & 97.9 & \multicolumn{1}{l|}{84.8} & \multicolumn{1}{l|}{77.4} & 38.9 \\ \cline{2-8} 
                     & Ghani          & \multicolumn{1}{l|}{77.3}          & \multicolumn{1}{l|}{90.0} & 94.9 & \multicolumn{1}{l|}{79.1} & \multicolumn{1}{l|}{62.9} & 31.8 \\ \hline
\multirow{2}{*}{5} &
  Modern &
  \multicolumn{1}{l|}{84.0} &
  \multicolumn{1}{l|}{95.1} &
  98.1 &
  \multicolumn{1}{l|}{85.3} &
  \multicolumn{1}{l|}{75.6} & 40.0
   \\ \cline{2-8} 
                     & Ghani & \multicolumn{1}{l|}{77.6} & \multicolumn{1}{l|}{90.1} & 94.9 & \multicolumn{1}{l|}{79.5} & \multicolumn{1}{l|}{61.6} & 31.7 \\ \hline
\multirow{2}{*}{3} & Modern         & \multicolumn{1}{l|}{82.8}          & \multicolumn{1}{l|}{94.4} & 97.6 & \multicolumn{1}{l|}{84.4} & \multicolumn{1}{l|}{71.8} & 42.1 \\ \cline{2-8} 
                     & Ghani          & \multicolumn{1}{l|}{77.4}          & \multicolumn{1}{l|}{90.0} & 94.8 & \multicolumn{1}{l|}{79.4} & \multicolumn{1}{l|}{59.7} & 32.1 \\ \hline
\end{tabular}
\caption{\label{results-windows}Baselines - evaluation of ArabGlossBERT on two sense inventories, with different context windows and sense orderings.}
\end{table}

Table \ref{results-windows} presents further evaluation of ArabGlossBERT, which illustrates the following:  
(i) using Modern is better than using Ghani in all experiments. This might be because of the better quality of Modern glosses (refer to IAA in Section \ref{sec:iaa});
(ii) While window 11 and 5 have the highest WSD accuracy, the use of context windows does not make major difference (only 1.4\% for Modern and 0.6\% for Ghani); 
 (iii) the ranking of the intended sense among the top 1, 2, and 3 senses illustrates a consistent and reasonable increase in the WSD accuracy; and (iv) when evaluating the model accuracy for noun and verb, the accuracy of nouns is about 8.5\% better than verbs for Modern, which might be because verbs are typically more ambiguous \cite{MJK23}. The WSD accuracy for functional words is very low with both lexicons. This is because functional words are highly polysemous and their glosses describe their functions rather than semantics.

\section{Conclusion}
\label{sec:conclusion}
We presented SALMA, the first sense-annotated Arabic corpus. The novelty of SALMA lies in utilizing two sense inventories and named entity annotations. In addition, instead of linking a word to one intended sense, we scored all semantically related senses of each token in the corpus. The quality of the annotations was assessed using various inter-annotator agreement metrics (Kappa, LWK, QWK, MAE, and RSME). To compute a WSD baseline using our corpus, we proposed to build an end-to-end WSD system using TSV, and evaluated this system using three different TSV models. The full corpus, annotations, and the tool, are open source and publicly available on GitHub.

\section{Limitations and Future Work}
\label{sec:limits-future}
Although Modern provides a better quality of glosses compared with the Ghani, some of Modern's glosses are referrals, i.e., referred to another related lemma. At this stage, we annotated these referrals as senses. Nevertheless, in order to use the Modern as a general sense inventory, these referrals need to be treated differently. We plan to replace all referral glosses with the senses they refer to, which can be done semi-automatically. For missing lemmas in Modern, we plan to map between the lemmas in both lexicons and then import missing lemmas and their senses from Ghani to Modern. In this way, we expect to have a richer Arabic sense inventory. Additionally, our sense annotations are limited to the senses of a single-word lemma. We plan to annotate the corpus with multiword expressions \cite{JZAA18}. Furthermore, the corpus we presented in this article is limited to MSA. To extend this corpus with dialectal text, plan to sense-annotate portions of the available corpora Curras \cite{EJHZ22,JHRAZ17}, Baladi \cite{EJHZ22}, Nabra \cite{ANMFTM23} and Lisan \cite{JZHNW23}.



\section*{Acknowledgment}
\label{sec:ack}
We would like to thank Shimaa Hamayel, Tamara Qaimari, Raghad Aburahma, Hiba Zayed, and Rwaa Assi for helping us in the corpus annotation.

\bibliography{emnlp2023}
\bibliographystyle{acl_natbib}

\end{document}